\documentclass[10pt, a4paper]{article}
\usepackage{lrec2026}

\setcitestyle{numbers,square}

\usepackage{graphicx}
\usepackage{amsmath}
\usepackage{amssymb}
\usepackage[ruled,vlined]{algorithm2e}
\usepackage{enumitem}
\usepackage{url}

\title{DiffuSyn Bench: Evaluating Vision Language Models on Real-World Complexities\\with Diffusion Generated Synthetic Benchmarks}

\name{Haokun Zhou, Yipeng Hong}
\address{Imperial College London\\
        Shanghai University\\
         \texttt{haokun.zhou25@imperial.ac.uk, 
         hongyipeng@shu.edu.cn}}

\abstract{
This study assesses the ability of Large Vision-Language Models (LVLMs) to differentiate between AI-generated and human-generated images and introduces a new automated benchmark construction method for this evaluation. The experiment compared common LVLMs with human participants using a mixed dataset of AI- and human-created images. Results showed that LVLMs could distinguish between the image types to some extent but exhibited a bias toward classifying images as human-generated and performed significantly worse than humans. To build on these findings, we developed an automated benchmark construction process using AI. This process involved topic retrieval, narrative script generation, error embedding, and image generation, creating a diverse set of text-image pairs with intentional errors. We validated our method by constructing two comparable benchmarks. This study highlights the strengths and weaknesses of LVLMs in real-world understanding and advances benchmark construction techniques, providing a scalable and automatic approach for AI model evaluation.
\\ \newline \Keywords{Vision-language model, Benchmark, Synthetic data, Latent diffusion model}
}

\begin{document}

\maketitleabstract

\section{Introduction}
\begin{figure}[!ht]
\centering
\includegraphics[width=\columnwidth]{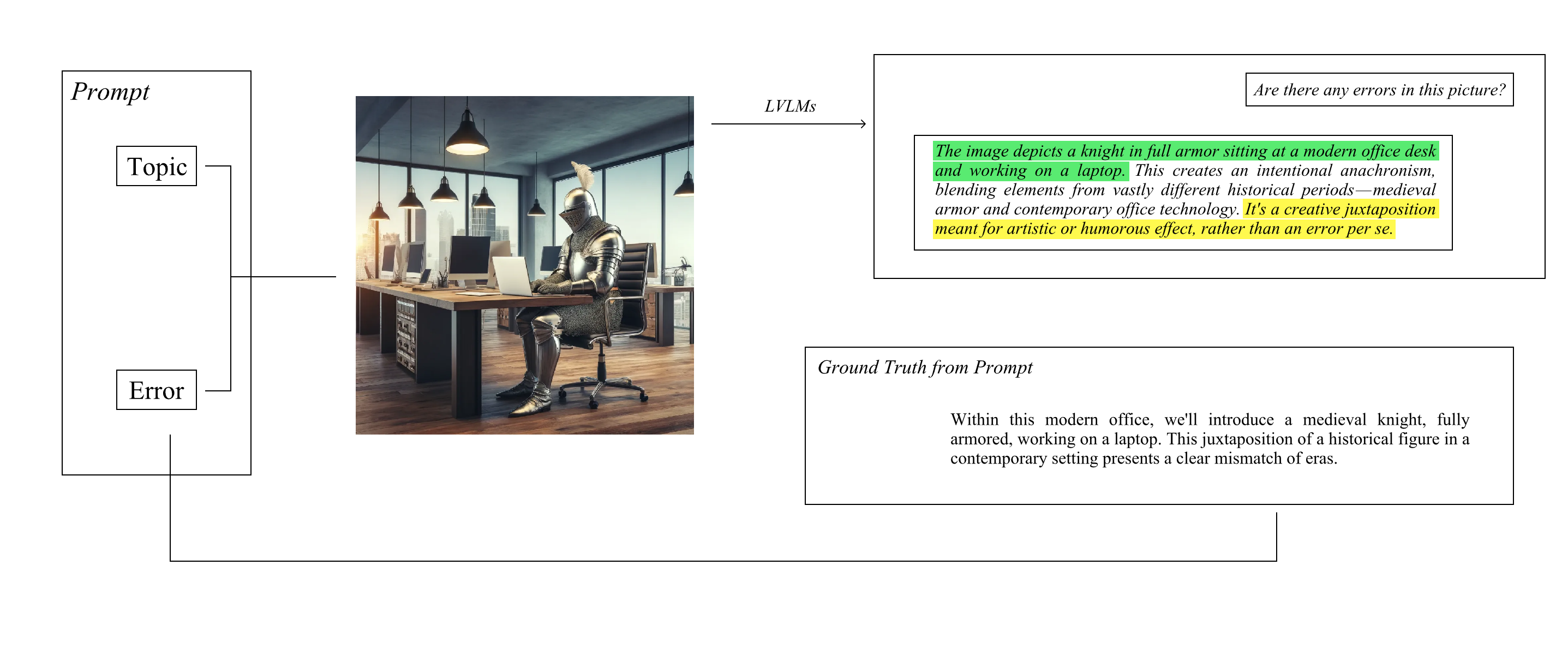}
\caption{A brief framework for benchmark generation and a specific example from our benchmark, showing a GPT-4V response. The example image depicts a modern office with a medieval knight using a laptop, highlighting intentional anachronism.}
\label{fig6}
\end{figure}

In recent years, remarkable advances in deep learning within the fields of computer vision and natural language processing have propelled LVLMs to the forefront as powerful tools for understanding and interpreting visual and textual data. Despite their impressive capabilities, the extent to which these models can accurately comprehend and analyze real-world scenarios remains a subject of ongoing investigation. This paper embarks on a twofold exploration: firstly, it assesses the capacity of LVLMs to differentiate between AI-generated and human-originated images—a task that probes their ability to interpret and understand nuanced visual information. Secondly, it extends these insights to the systematic development of a methodology for the automatic generation of benchmarks designed to evaluate this ability.

The first part of our study focuses on a rigorous examination of LVLMs' performance in identifying anomalies and errors within images, utilizing a dataset comprising both AI-generated and human-created visuals. This evaluation is critical for understanding the models' proficiency in real-world understanding, as well as their ability to discern subtle differences in image quality and content.

Building upon these findings, the second part of the paper delves into the development of an automatic benchmark construction methodology. This approach uses AI techniques to generate and validate benchmarks, providing a scalable and efficient means of testing LVLMs' capabilities. By integrating automated processes, we aim to create robust and reliable benchmarks that can be utilized to continuously assess and improve the performance of LVLMs in various real-world tasks.

\paragraph{Contributions summary.} We (i) quantify a human--LVLM perception gap on image-origin judgments; (ii) introduce a scalable automated pipeline (topic retrieval $\rightarrow$ narrative $\rightarrow$ error embedding $\rightarrow$ image generation) with LLM-as-a-Judge (LAJ) based validation and scoring; (iii) define a three-way taxonomy (temporal, biological, logical) and construct 848 text--image pairs; and (iv) externally validate the approach with comparable benchmarks while preserving model ordering.

\subsection{Related Work \& Design Rationale}
Recent multimodal benchmarks probe diverse capabilities (e.g., MM‑Vet, HallusionBench, VisIT‑Bench, VASR), while graphics‑based pipelines such as PUG emphasize photorealistic scene construction. In contrast, our framework targets criteria‑aware, fine‑grained inconsistencies (temporal/biological/logical) and uses an LLM‑as‑a‑Judge to score open‑ended descriptions, yielding a scalable, human‑aligned evaluation protocol \citep{ref_mmvet,ref_hallusionbench,ref_visitbench,ref_vasr,ref_pug}.

\paragraph{Comparison with engine rendered synthetic benchmarks.}
Engine-rendered pipelines such as PUG \citep{ref_pug} generate photorealistic, semantically controllable scenes with precise control over lighting, camera, and object placement, and they provide perfect pixel-level annotations. This makes them well suited for evaluating capabilities within plausible physical worlds. However, for our goal, probing LVLMs’ sensitivity to real-world inconsistencies (biological impossibilities, cross-period anachronisms, violations of object affordances), engine pipelines are constrained by asset availability and built-in physics. Creating counterfactual scenes typically requires bespoke asset modeling/rigging or shader programming, which limits diversity and scale.

DiffuSyn takes a complementary approach: diffusion-based text-to-image generation composes novel and sometimes impossible combinations directly from language, enabling scalable synthesis of diverse out-of-distribution cases. Prompt validation ensures the intended inconsistency is realized in the image (Fig.~\ref{fig3}), and LLM-as-a-Judge (LAJ) scoring provides criteria-aware, human-aligned evaluation (Sec.~\ref{sec:laj-why}). In practice, DiffuSyn is preferable when evaluating robustness to inconsistencies that are impractical to author in graphics engines, while PUG-like pipelines excel when fine-grained parametric control and perfect annotations are required.

\section{Methods: The DiffuSyn Benchmark Framework}
\label{sec:methods}
To rigorously evaluate the capacity of LVLMs to comprehend nuanced real-world complexities—spanning physical laws, biological constraints, and temporal coherence—we introduce \emph{DiffuSyn Bench}. This benchmark is characterized by a novel, automated computational pipeline that synthesizes photorealistic imagery embedded with controlled, context-specific errors. We first quantify the limitations motivating this approach in Section~\ref{sec:exp1}.

\subsection{Automated Benchmark Construction and Evaluation}
We propose a fully automated pipeline for generating the DiffuSyn Bench. This methodology facilitates the scalable creation of diverse text--image pairs featuring intentional, contextually embedded errors, minimizing human labor and mitigating risks of data contamination. To keep evaluation human-aligned and reproducible, we follow an LLM-as-a-Judge (LAJ) paradigm for both prompt validation and model scoring; detailed justification and reliability controls appear in Sec.~\ref{sec:laj-why}.

\subsubsection{LAJ instead of embedding based approach}
\label{sec:laj-why}
Our task requires fine-grained, criteria-aware reasoning about specific error types (temporal, biological, logical). Single-vector similarity metrics (e.g., CLIPScore, BERTScore) are effective for \emph{coarse} image--text alignment but are known to (i) conflate topical relatedness with factual correctness, (ii) penalize valid paraphrases, and (iii) miss localized or plausibility-related errors: the core phenomena we target \citep{ref_clipscore, ref_bertscore, ref_metrics_robustness}. 

By contrast, LAJ allows an evaluator LLM to apply an explicit rubric, decompose the task, and return interpretable judgments that better track human preferences on open-ended outputs. Large-scale studies show that GPT-4 judges achieve high agreement with human raters on multi-turn evaluations (MT-Bench / Chatbot Arena), and that rubric-prompted GPT-4 evaluators (G-Eval) yield higher human correlation than prior automatic metrics on summarization and dialogue \citep{ref_laj_mtbench, ref_geval}.

For vision--language tasks specifically, LLM-based approaches have surpassed embedding methods on faithfulness and error sensitivity. \citet{ref_tifa} evaluate text-to-image alignment by generating fine-grained questions with an LLM and checking answers with VQA, reporting stronger correlation with human judgments and improved sensitivity over CLIP-style metrics. Multimodal diagnostic suites such as \citet{ref_hallusionbench} also adopt GPT-4/GPT-4V assisted judging and show that LAJ-based scoring tracks human assessment while diagnosing multimodal hallucinations. These results support our choice of a \emph{criteria-aware LLM judge} over global embedding similarity to detect nuanced, real-world inconsistencies.

\begin{figure}[!ht]
\centering
\includegraphics[width=\columnwidth]{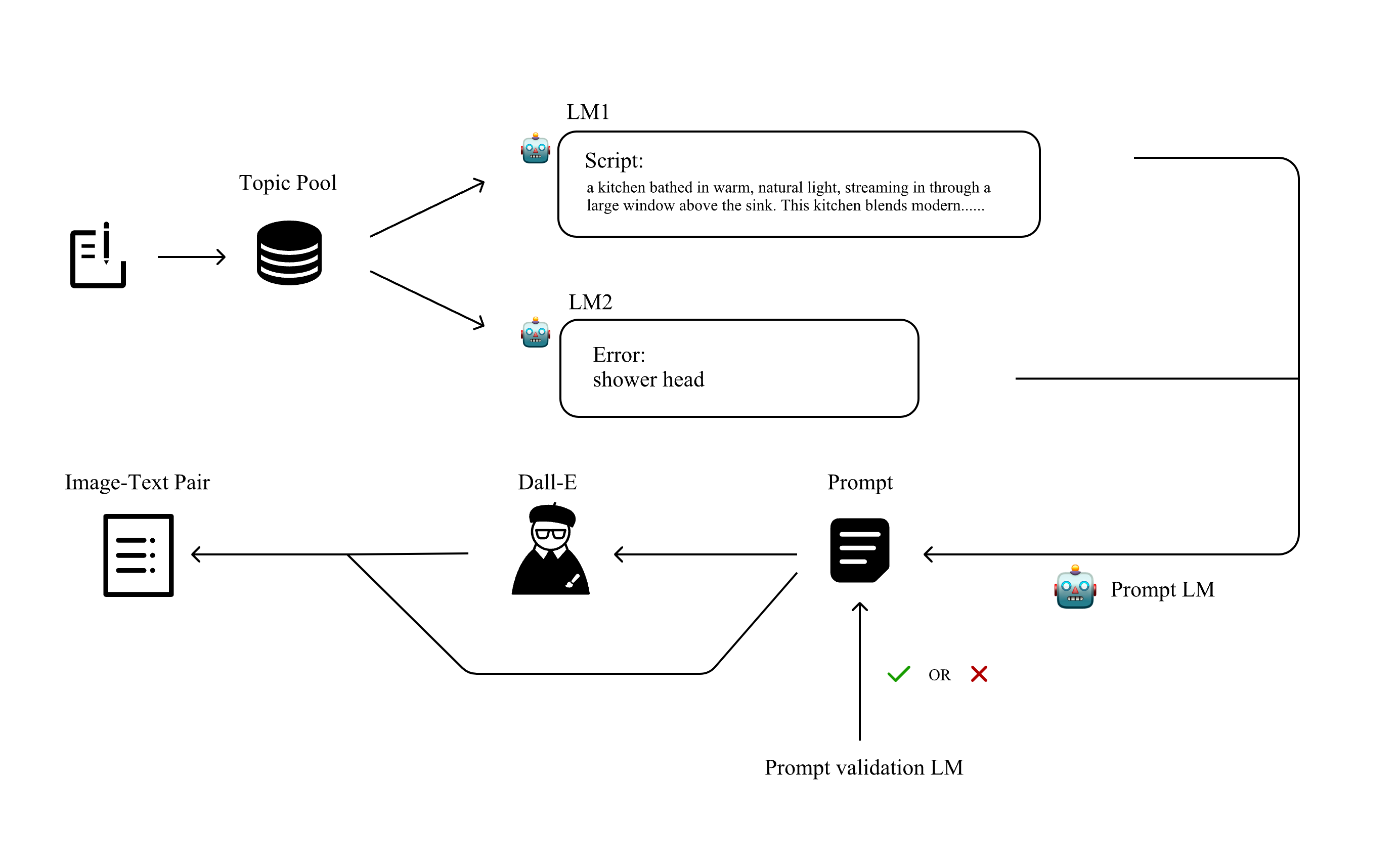}
\caption{Automated framework to generate diffusion based synthetic evaluations.}
\label{fig3}
\end{figure}

\paragraph{Our usage and reliability controls.}
We invoke LAJ twice in our pipeline: \textbf{(i)} Prompt validation: a judge screens candidate prompts for visualizability and error salience before image generation (mitigating failure cases where T2I cannot manifest the intended error; see Fig.~\ref{fig3}), and \textbf{(ii)} Output scoring: a judge compares each LVLM response to the reference error description via a task-specific rubric and assigns a 0--10 score for accuracy and completeness. To reduce judge variance and biases, we fix temperature to \(0\), standardize evaluation prompts, and aggregate \(K{=}3\) independent judge runs by mean score or majority vote.

\subsubsection{Pipeline Architecture and Rationale}
\label{sec:pipeline}
The pipeline employs a multi-agent architecture, utilizing the synergistic interaction of four distinct Language Model instances and a Text-to-Image model. This design is crucial for overcoming inherent limitations in T2I generation, such as the lack of precise output control and difficulties in complex concept composition \citep{ref_photorealistic, ref_structured}. Our architecture addresses this through a structured division of labor.

\subsubsection{Stage 1: Contextualization and Error Definition}
The process initiates with topic retrieval, where a conceptual theme (e.g., ``Victorian Laboratory,'' ``Coral Reef'') is sampled from a predefined topic pool \(P\). This theme is simultaneously input to two parallel LM agents:
\begin{itemize}[leftmargin=1.25em]
  \item \textbf{LM1 (Narrative Generation):} Generates a detailed, contextually rich narrative script \(S\) based on the topic, establishing the scene's foundational elements.
  \item \textbf{LM2 (Error Generation):} Concurrently devises a specific inconsistency \(E\) relevant to the topic, designed to violate a real-world constraint.
\end{itemize}

\subsubsection{Stage 2: Prompt Synthesis and LLM-based Validation}
A third Prompt LM synthesizes a cohesive T2I prompt \(P_{\mathrm{T2I}}\) by integrating the narrative script \(S\) and the error \(E\).

Crucially, we introduce an innovative Prompt Validation stage. Many conceptually sound errors are difficult for current T2I models to visualize accurately (e.g., complex physical paradoxes like ``the shadow facing the sun''). Attempting to generate these often results in noise, where the image fails to manifest the intended error, thereby degrading benchmark quality.

We employ a dedicated validation LM to assess the synthesized prompt \(P_{\mathrm{T2I}}\) for its visual feasibility. This agent acts as a discriminator, predicting whether the T2I model can successfully render the embedded error with sufficient prominence. Only validated prompts proceed to image generation.

This validation mechanism significantly enhances the pipeline's efficacy. It reduced the generation failure rate from 28.1\% (in a single-LM baseline) to 5.8\%. Moreover, it enforces diversity by mitigating the bias towards easily rendered scenarios. In the baseline approach, common scenes (kitchens, offices) dominated \(>90\%\) of the dataset; our method reduced the prevalence of any single scene to \(<5\%\).

\subsubsection{Stage 3: Image Generation}
The validated prompt \(P_{\mathrm{T2I}}\) is input to the DALL-E~3 T2I model, which generates the final image \(I\), visually manifesting the intended error within the specified context. The resulting pair \(\big(P_{\mathrm{T2I}}, I)\) forms an entry in the DiffuSyn Bench.

\subsection{Benchmark Taxonomy}
The DiffuSyn Bench dataset \(D_{S}\) is organized around a taxonomy designed to probe distinct facets of real-world understanding. We categorize the embedded errors into three orthogonal types:
\begin{description}[leftmargin=1.5em]
  \item[Biological Errors:] Violations of anatomical structures (e.g., incorrect number of limbs).
  \item[Temporal Errors:] Anachronisms and the juxtaposition of objects or styles from disparate historical periods without logical coherence (e.g., a medieval knight using a laptop, as shown in Fig.~\ref{fig6}).
  \item[Logical Inconsistencies:] Violations of physical functionality (e.g., a teapot made of wool) or spatial incongruities (e.g., doors opening into impossible spaces).
\end{description}

The finalized DiffuSyn Bench comprises 848 text--image pairs, distributed across the Temporal (287), Biological (289), and Logical (272) categories.

\subsection{Evaluation Protocol}
The LVLMs were presented with each image in the dataset and tasked with identifying any anomalies or errors indicative of AI generation. Directly asking if the image is AI-generated or contains errors may result in a refusal to answer or affect the result due to the alignment of the LVLM. Therefore, following their analysis, the LVLMs provided a descriptive response for each image.

The responses from the LVLMs were then processed through GPT-3.5-Turbo-1106, which interpreted the descriptive answers and converted them into binary outcomes: `AI-generated' or `Human-generated'. This LAJ step standardizes results for statistical comparison; for the motivation, design choices, and bias/reliability controls behind our LAJ setup (including rubric design and aggregation), see Sec.~\ref{sec:laj-why}.

\section{Experiment 1: Perception Gap (Human vs.\ LVLM)}
\label{sec:exp1}

\subsection{Dataset Curation and Experimental Conditions}
We assembled a dataset \(D_{M}\) comprising \(N=2000\) images, equally partitioned into AI-generated \((D_{\mathrm{AI}})\) and human-originated \((D_{\mathrm{H}})\) subsets. To ensure diversity in generative artifacts, \(D_{\mathrm{AI}}\) included outputs from multiple diffusion architectures: Stable Diffusion~1.5, Stable Diffusion~2.1 (sourced from the DiffusionDB repository), and DALL-E~3 . The \(D_{\mathrm{H}}\) subset was curated from diverse public collections to ensure broad coverage of styles and subjects. All images were standardized to a resolution of \(512\times512\) pixels.

We evaluated a selection of prominent LVLMs, including GPT-4V (specifically the GPT-4-1106-Vision-Preview), Fuyu-8B, CogVLM, Qwen-VL, and LLaVA-1.5-13B. To ensure reproducibility and minimize stochasticity in the outputs, the temperature parameter for all model inferences was fixed at \(T=0\).

\subsection{Evaluation Procedure and Interpretation}
Directly querying LVLMs about the origin of an image (AI or human) often encounters resistance due to model alignment protocols, leading to refusals or biased responses. To circumvent this, we employed an indirect prompting strategy. Models were prompted to analyze and describe the image quality and identify any potential anomalies or inconsistencies.

The resulting descriptive text outputs were non-trivial to evaluate automatically. We utilized GPT-3.5-Turbo-1106 as a standardized interpreter to convert these nuanced descriptions into a binary classification (AI-generated or Human-generated). This conversion relies on the interpreter identifying keywords and patterns indicative of synthetic artifacts within the LVLM's description.

A human baseline was established by recruiting 50 participants naive to computer vision or NLP research. Participants were evaluated on a randomized subset of 200 images following a brief 5-shot familiarization session.

For all binary metrics, we define the positive class as “AI‑generated”. We aggregate the interpreter’s \(K=3\) runs by majority vote.

\subsection{Foundational Findings}
The analysis revealed a significant performance differential (detailed in Section~\ref{sec:quantitative}). Human observers demonstrated high accuracy (88.45\%, F1-score: 88.1\%). In contrast, the highest-performing LVLM, Fuyu-8B, achieved only 66.1\% accuracy. Notably, the confusion matrices (Fig.~\ref{fig1}) indicated a pronounced bias toward classifying images as human-originated (a high False Negative rate regarding AI detection). Although \(\chi^2\) tests confirmed that most models (excluding LLaVA-1.5-13B) performed better than chance, the overall results suggest that LVLMs lack the robust real-world understanding necessary to reliably detect subtle deviations from reality.

This perception gap motivates the development of a benchmark that moves beyond arbitrary generative artifacts, focusing instead on systematically probing an LVLM's understanding of complex, real-world constraints. It also motivates Section~\ref{sec:exp2}.

\subsection{Observations and Analysis}
\begin{figure}[!ht]
\centering
\includegraphics[width=\columnwidth]{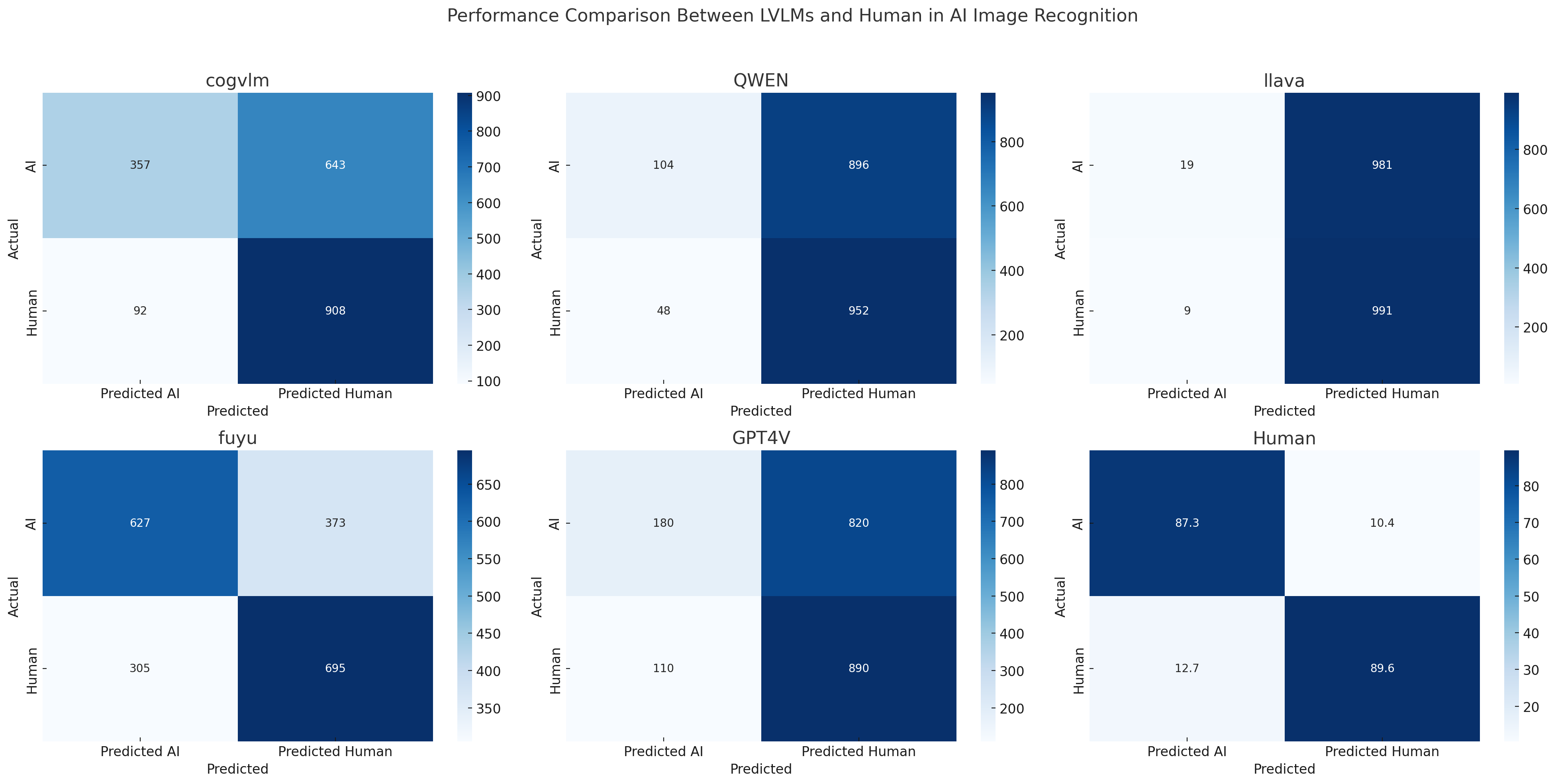}
\caption{Performance comparison between LVLMs and humans in distinguishing AI-generated vs.\ human-generated images (six heatmaps, one per system).}
\label{fig1}
\end{figure}

We analyze performance with confusion matrices (TP/FP/TN/FN). LVLMs show a strong bias toward the “human” label, inflating FN and depressing TP. This pattern is consistent with limited sensitivity to subtle textural/physical artifacts characteristic of synthetic imagery and, in part, alignment‑driven reluctance to assert AI origin. Humans exhibit the complementary error—flagging artifact‑free photographs as AI—raising FP. Inspecting model outputs, GPT‑4V in particular often rejects AI attributions even when diagnostic artifacts are present or, at times, after having described them. The resulting asymmetry indicates a systemic alignment bias that suppresses positive calls and hampers accurate AI detection (see Fig.~\ref{fig1}).

\subsection{Quantitative Analysis}
\label{sec:quantitative}
We report accuracy and F1 (positive = AI-generated). Formally,
\[
\mathrm{F1} = \frac{2\,\mathrm{TP}}{2\,\mathrm{TP}+\mathrm{FP}+\mathrm{FN}}.
\]

Humans reached 88.45\% accuracy. If pooled over $n=2000$ images, the 95\% binomial CI is [87.05, 89.85]\%; if aggregated over 10\,000 individual decisions ($50\times 200$), the 95\% CI is [87.82, 89.08]\%. Fuyu-8B achieved 66.1\% accuracy ($n=2000$; 95\% CI [64.03, 68.17]\%) and LLaVA-1.5-13B 50.5\% ($n=2000$; 95\% CI [48.31, 52.69]\%). Using a binomial test vs. 50\% (chance), Fuyu-8B is above chance ($z=14.40$, two-sided $p \ll 10^{-3}$), while LLaVA-1.5-13B is not ($z=0.447$, $p=0.655$).

Reported F1 scores (positive = AI) are: CogVLM 48.9, Qwen-VL 18.1, LLaVA-1.5-13B 3.7, Fuyu-8B 64.9, GPT-4V 28.0—all well below human 88.1. Overall, LVLMs show a bias toward the ``human'' label, inflating FN and depressing TP (see Fig.~3).

\begin{figure}[!ht]
\centering
\includegraphics[width=\columnwidth]{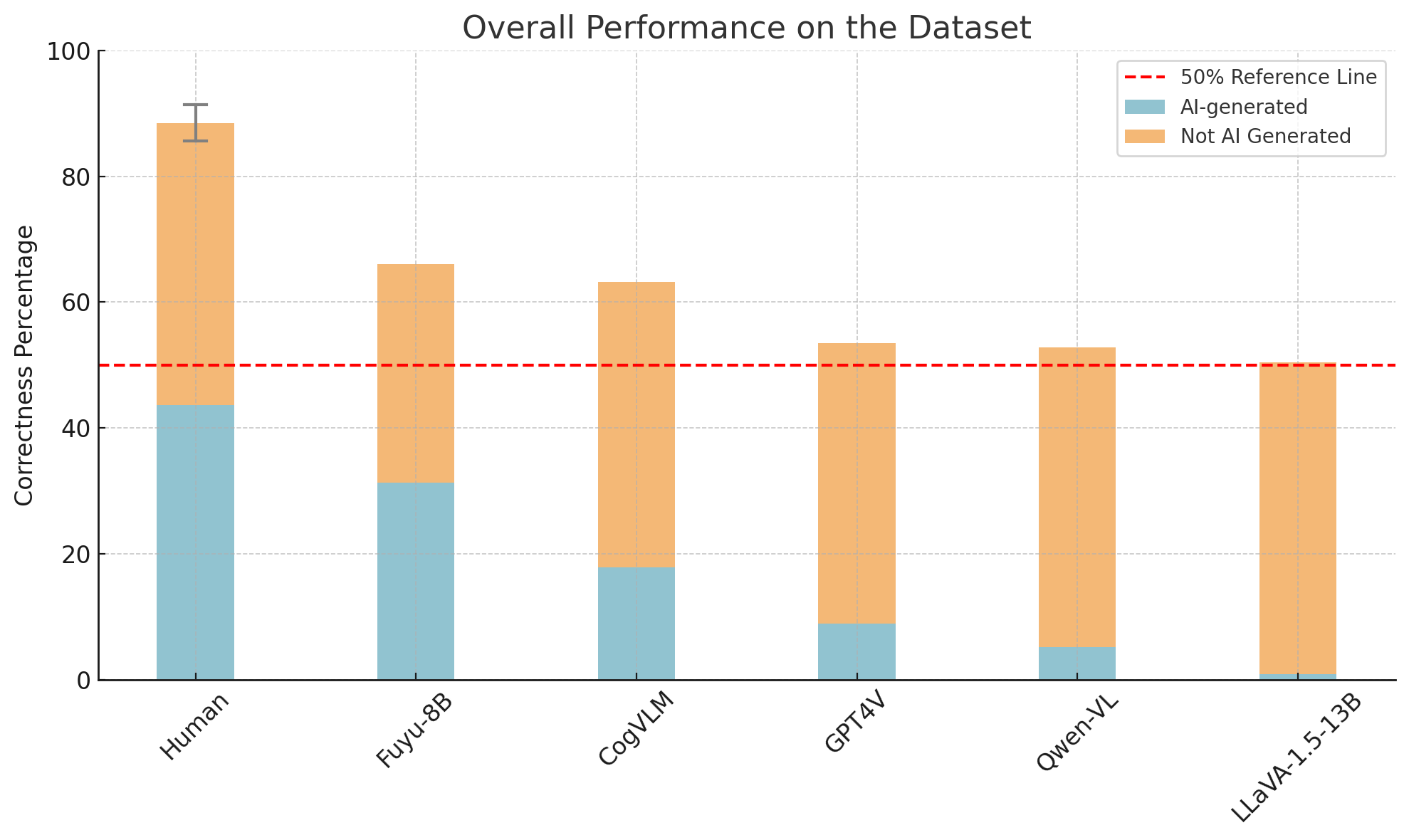}
\caption{Overall correctness of humans and models on AI- vs.\ non-AI-generated images. The red dashed line (50\%) marks random-guessing.}
\label{fig2}
\end{figure}

Chi-square tests for independence corroborate LVLMs' capacity to identify AI-generated content beyond chance, except for the LLaVA-1.5-13B model. This statistical significance underscores LVLMs' innate ability to detect errors indicative of AI-generated images, notwithstanding their inherent limitations. (see Fig.~\ref{fig2})

\section{Experiment 2: Results on DiffuSyn Bench}
\label{sec:exp2}

\subsection{Task and Scoring Protocol}
We evaluated four LVLMs (CogVLM, GPT-4V, LLaVA, and Qwen) on DiffuSyn Bench. For each image, the model described the embedded error; an LLM-as-a-judge assigned a score from 0 to 10 by comparing the description to the ground-truth error statement. Scores were aggregated by error type. The finalized benchmark contains 848 text--image pairs distributed across temporal (n=287), biological (n=289), and logical (n=272) categories. The pipeline, taxonomy, and dataset statistics are shown in Figs.~\ref{fig6} and \ref{fig3}.

When using LLMs as judges, prior peer-reviewed work suggests they can correlate well with humans in some settings while diverging in others. For text-only evaluation, GPT-4–based judges achieve strong correlations with human assessments on summarization and dialogue, though biases remain \cite{ref_geval}. In the multimodal setting, an ICML study shows that MLLMs-as-judges align more closely with human preferences in pairwise comparison than in scalar scoring and batch ranking, where notable divergences and biases appear \cite{chen2024mllmjudge}.

\subsection{Aggregate Performance by Error Type}
\begin{figure}[!ht]
\centering
\includegraphics[width=\columnwidth]{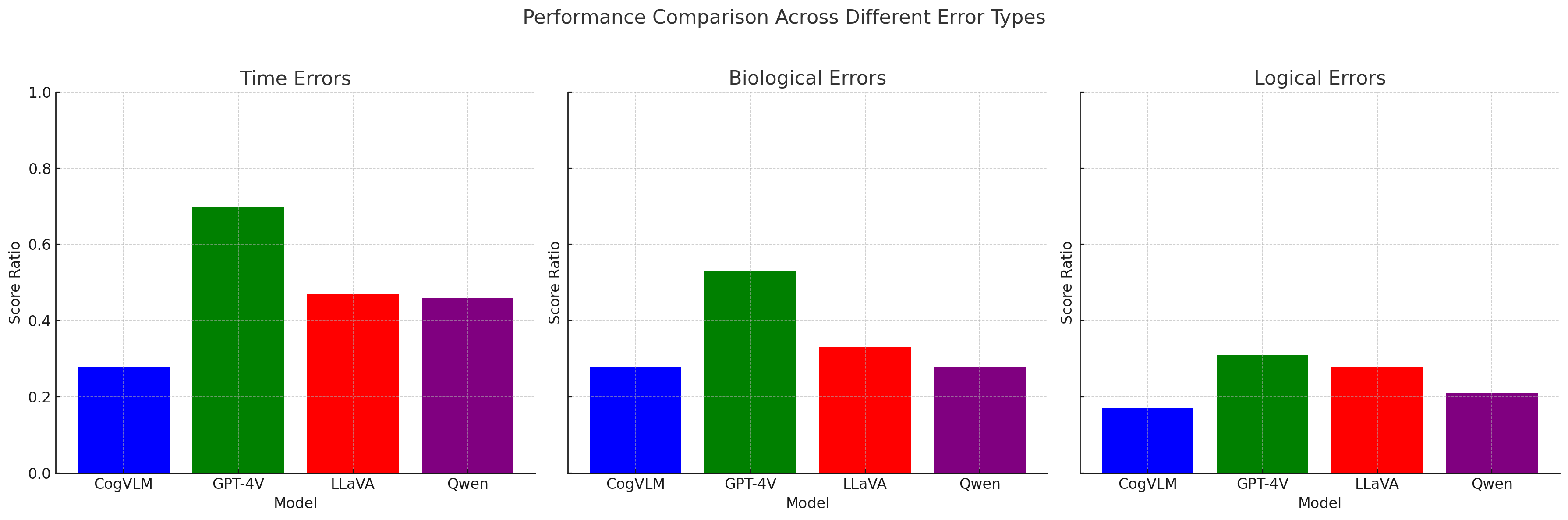}
\caption{Performance of four models across three error types. GPT-4V attains the highest totals across temporal, biological, and logical categories.}
\label{fig4}
\end{figure}

Figure~\ref{fig4} reports totals by category. GPT-4V attains the highest scores for temporal (2031; LLaVA: 1347), biological (1268; LLaVA: 796; Qwen: 657), and logical errors (177; LLaVA: 164; Qwen: 121). All models degrade substantially outside the temporal category, and the manuscript attributes GPT-4V’s comparatively low absolute logical totals partly to alignment-induced caution: in uncertain cases it refrains from making explicit error claims, which depresses scores when decisive identification is expected.

\subsection{Miscategorization Analysis}
To analyze difficulty rather than only magnitude, we normalized by category size. For GPT-4V, the average points per item are approximately 7.08 (temporal; 2031/287), 4.39 (biological; 1268/289), and 0.65 (logical; 177/272). For LLaVA, the corresponding averages are approximately 4.69, 2.75, and 0.60 (1347/287; 796/289; 164/272). These gradients indicate that temporal inconsistencies are least often missed or misattributed, biological inconsistencies are more frequently described imprecisely, and logical inconsistencies are most often overlooked or reframed as stylistic oddities. This interpretation is consistent with the manuscript’s observation that current LVLMs struggle when fine-grained details must be reconciled with real-world constraints.

\begin{figure}[!ht]
\centering
\includegraphics[width=\columnwidth]{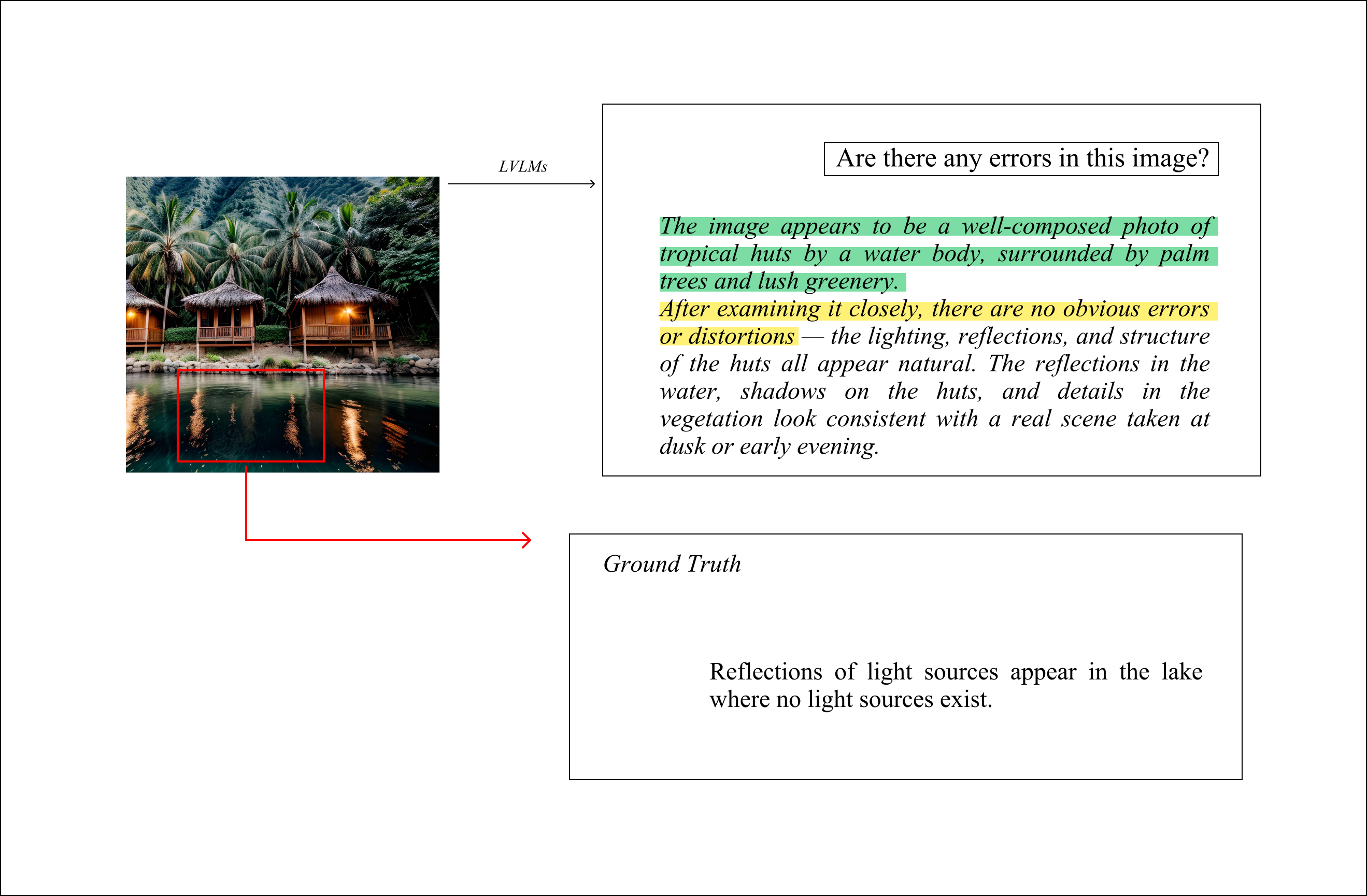}
\caption{A specific example from our benchmark, showing a GPT-4V response, demonstrate limited abilities of LVLMs to reasoning through real-world physics and understand the problem.}
\label{fig7}
\end{figure}

This pattern is also consonant with long-standing evidence that performance drops when the diagnostic region is small or occluded. The COCO benchmark formalized evaluation by object size (AP$_S$, AP$_M$, AP$_L$), with small objects historically exhibiting the lowest accuracy, highlighting the challenge of attending to and reasoning over small regions in complex scenes \cite{lin2014coco}. To see how LVLMs are struggled at this area, please refer to Figure~\ref{fig7}.

\subsection{Why Do These Patterns Emerge?}
A first mechanism is error salience and spatial extent. Temporal violations in our benchmark are often global and scene-level; by contrast, many biological and logical violations occupy small, local regions. When discrepant areas are small or partially occluded, models under-attend to the critical evidence and produce underspecified descriptions. This account matches both the normalized results above and the cross-benchmark experiment below.

A second mechanism is generation-side controllability and compositionality. Publicly available diffusion outputs frequently display attribute-binding failures or concept neglect unless additional control is imposed. Peer-reviewed studies document these issues and propose remedies: attention-based guidance that prevents catastrophic neglect of prompt tokens \cite{chefer2023attend}, explicit spatial conditioning via ControlNet \cite{zhang2023controlnet}, structured composition of diffusion sub-models \cite{liu2022composable}, and compositional stress tests revealing persistent failures in color/shape binding and object relations \cite{huang2023t2icomp}. These results support our observation that when prompts are imperfectly controllable, errors tend to be subtle and local, making them harder for LVLMs to identify decisively.

A third mechanism is affordance and physical reasoning. Many logical items require judging whether an object could function as depicted. Recent CVPR work shows that even foundation-model–based approaches need specialized treatment to generalize fine-grained affordances, underscoring why logical contradictions with no overt pixel-level artifact are easily missed \cite{li2024ooal}.

\subsection{External Validation via Alternative Benchmark Constructions}
\begin{figure}[!ht]
\centering
\includegraphics[width=\columnwidth]{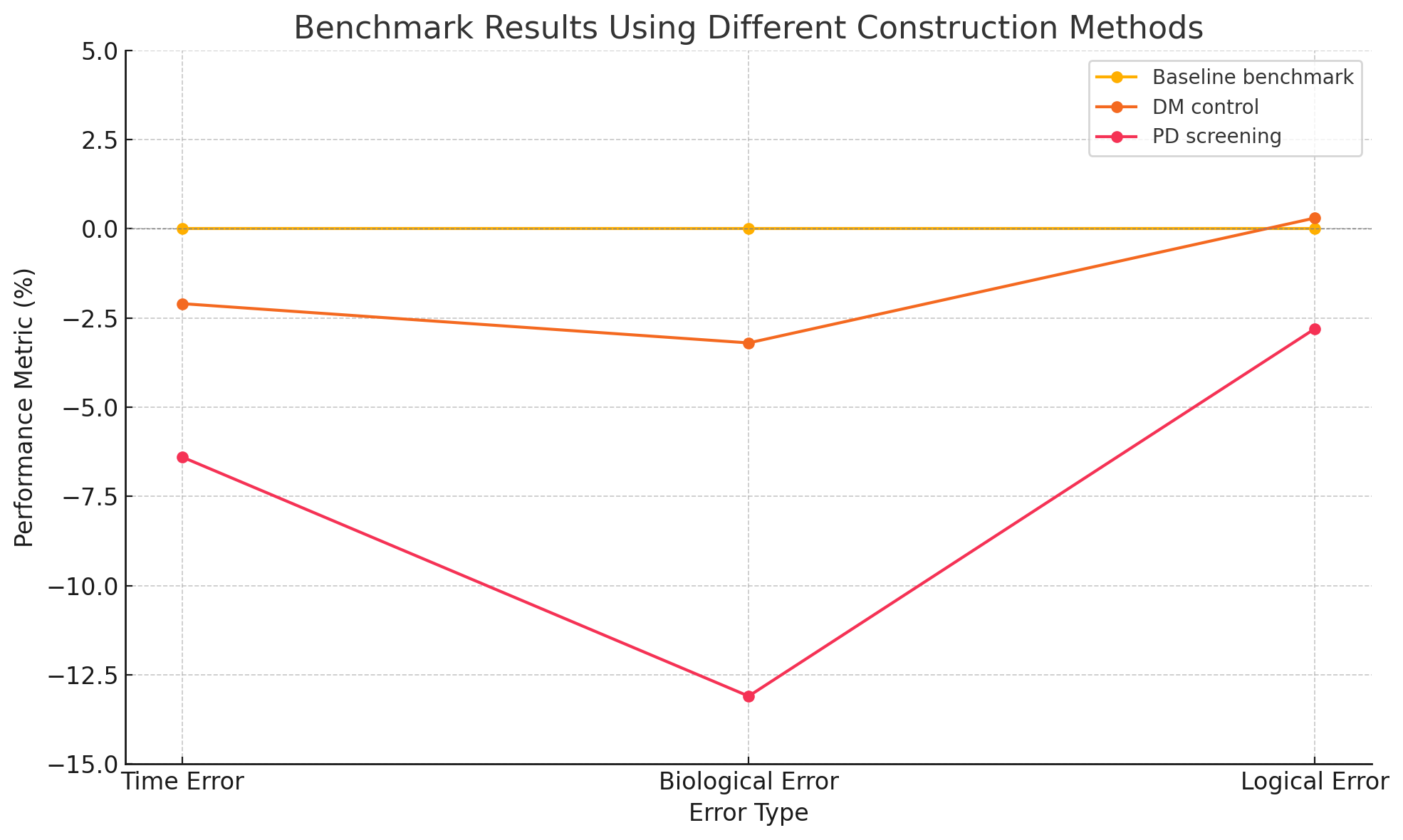}
\caption{Benchmark results using different construction methods. Yellow: original synthetic benchmark; orange: human-authored prompts; red: manually screened diffusion images.}
\label{fig5}
\end{figure}

To validate the construction method, we built two comparable benchmarks and re-evaluated the same LVLMs: a manually screened subset of a public diffusion dataset with obvious errors and manual error descriptions, and a benchmark generated from human-authored prompts containing errors using the same image generator as the main pipeline. As shown in Figure~\ref{fig5}, the human-authored prompts benchmark yields no material differences relative to the original synthetic set, whereas all models’ scores decrease on the manually screened set. The manuscript attributes this decrease to limited prompt controllability and the prevalence of errors arising from diffusion-model limitations rather than the intended prompt, which produce less obvious anomalies with smaller spatial extent. Despite lower absolute performance on the manually screened set, the model ranking is preserved. With four models, the induced ranks are identical, yielding Spearman’s $\rho = 1.00$. Because $n = 4$, Spearman’s $\rho$ takes discrete values in steps of $0.2$; an exact two-sided permutation test gives $p = 0.083$ (one-sided $p = 0.0417$). We therefore report $\rho$ without a significance claim and emphasize the qualitative preservation of ordering.

This controllability explanation is consistent with peer-reviewed evidence that improving text--image alignment and localized control (e.g., ControlNet, attention-guided generation, composable diffusion) is often necessary to realize specified concepts reliably; absent such control, subtle, small-area discrepancies are common and difficult to detect \cite{zhang2023controlnet,chefer2023attend,liu2022composable,huang2023t2icomp}.

\subsection{Implications}
The concentration of miscategorization in biological and especially logical violations indicates that evaluation should manipulate the size and salience of error regions more systematically and include scenes that require cross-object affordance reasoning. Our alternative-benchmark results further suggest that automated synthesis can deliver valid assessments at scale while remaining sensitive to difficulty when error regions are small or subtle. Methodologically, future iterations should report item-level difficulty indicators (e.g., error-region area) and incorporate controllability mechanisms during data generation to probe targeted error types with higher fidelity. \cite{huang2023t2icomp,li2024ooal}

\subsection{Localized Error Injection via Emerging Image Editing Models}
Recent image‑editing models (e.g., GPT‑4o, Qwen Image Edit, Gemini 2.5 Flash) enable fine‑grained inpainting while preserving non‑edited regions. We therefore introduced an editing branch that bypasses the Prompt‑LM → T2I step (Sec. 2.1.4–2.1.5) and directly inserts the target inconsistency into a clean seed image. Concretely, the LAJ‑validated description of the intended error is converted into an edit instruction + mask, and an editing model inpaints only the masked area, leaving the remainder of the image unchanged.

Under our manual curation protocol, this editing‑based construction constrained the perturbation to <10\% of image pixels and raised the success rate for generating logical errors from 41\% to 72\%, while also broadening the repertoire of attainable inconsistencies that is impossible for image model to generate direct from prompt previously. For example, a tree without shadow. Because the edited region is small and salient by design, this branch operationalizes the recommendations in §4.6 to control error‑region size and improves fidelity on challenging logical cases.

\section{Limitations}
While our methodology aims to leverage AI for generating a comprehensive and diverse set of benchmarks, inherent limitations exist in both text-to-image (T2I) models and large language models (LLMs). The T2I models sometimes struggle to produce erroneous images following specific instructions, and LLMs exhibit limited ability to understand this characteristic of T2I models and craft clear, precise instructions for T2I models. This can introduce noise into the synthetic data, leading to less pronounced or unobservable errors and weakening the accuracy of the benchmarks.

To address this issue, we plan to enhance the precision of generated data and further reduce noise. Our efforts include constructing more powerful LLM agents combined with state-of-the-art T2I models, which we believe will help alleviate these challenges. By improving the clarity and specificity of instructions given to T2I models, and by refining the error generation process, we aim to create benchmarks that are both highly accurate and representative of real-world complexities. This will ultimately strengthen the evaluation process and contribute to more robust and reliable performance assessments of LVLMs.

\section*{Conclusion}
This study introduced DiffuSyn Bench, a novel framework designed to rigorously evaluate the capacity of LVLMs to comprehend nuanced, real-world complexities. Our research systematically established two critical findings.

First, we quantified a substantial perception gap between humans and state-of-the-art LVLMs in distinguishing AI-generated from human-originated imagery. Human observers significantly outperformed the best-performing LVLM, revealing that LVLMs struggle to reliably detect subtle deviations from reality and exhibit a marked bias toward classifying images as human-generated.

Second, motivated by this gap, we developed and validated a fully automated pipeline for synthetic benchmark construction. The DiffuSyn methodology, integrating multi-agent LLM orchestration for narrative generation and controlled error embedding (Temporal, Biological, and Logical) with diffusion-based image generation, offers a scalable and contamination-resistant approach to evaluation. Utilizing this pipeline, we created a dataset of 848 text-image pairs. Evaluations on this benchmark, scored using a criteria-aware LAJ protocol, highlighted critical weaknesses in current models. Performance significantly degraded when models encountered localized, subtle inconsistencies, particularly Logical errors requiring reasoning about physical affordances and constraints.

The validity of the DiffuSyn framework was confirmed through external validation, which preserved model rankings across comparable benchmarks. By advancing evaluation from static datasets towards dynamic, automated synthetic generation, this study underscores the necessity of targeted benchmarking to uncover limitations in real-world understanding. Our findings highlight the critical need for developing LVLMs that move beyond superficial pattern recognition toward deeper physical and causal reasoning to bridge the substantial gap between human and machine perception.

\end{document}